# Building A Knowledge Graph to Enrich ChatGPT Responses in Manufacturing Service Discovery

Yunqing Li[a], Binil Starly[b]
[a]Edward P. Fitts Department of Industrial & Systems Engineering, North Carolina State University
[b]School of Manufacturing Systems and Networks, Arizona State University

**ABSTRACT**
Sourcing and identification of new manufacturing partners is crucial for manufacturing system integrators to enhance agility and reduce risk through supply chain diversification in the global economy. The advent of advanced large language models has captured significant interest, due to their ability to generate comprehensive and articulate responses across a wide range of knowledge domains. However, the system often falls short in accuracy and completeness when responding to domain-specific inquiries, particularly in areas like manufacturing service discovery. This research explores the potential of leveraging Knowledge Graphs in conjunction with ChatGPT to streamline the process for prospective clients in identifying small manufacturing enterprises. In this study, we propose a method that integrates bottom-up ontology with advanced machine learning models to develop a Manufacturing Service Knowledge Graph from an array of structured and unstructured data sources, including the digital footprints of small-scale manufacturers throughout North America. The Knowledge Graph and the learned graph embedding vectors are leveraged to tackle intricate queries within the digital supply chain network, responding with enhanced reliability and greater interpretability. The approach highlighted is scalable to millions of entities that can be distributed to form a global Manufacturing Service Knowledge Network Graph that can potentially interconnect multiple types of Knowledge Graphs that span industry sectors, geopolitical boundaries, and business domains. The dataset developed for this study, now publicly accessible, encompasses more than 13,000 manufacturers' weblinks, manufacturing services, certifications, and location entity types.

------------------------------------------------------------------------------------------------------------

## 1. Introduction

The current global crisis has demonstrated the need for a shift towards a supply chain that is resilient, distributed, and digitally enabled. Recent worldwide crises such as the COVID-19 pandemic, workforce shortage, and logistics delays have strongly impacted supply chains across all major industrial sectors. This impact is significant, especially for small manufacturers that account for more than 75% of all US manufacturing business [1]. Generally, these companies spend a portion of their revenue on marketing their technical capabilities to promote their business and extensively rely on human network relationships. The global lockdown and extended travel restrictions during the pandemic prevent manufacturers from making those connections to help identify future business opportunities. With limited network connections and word-of-mouth marketing, we asked the question – *"how can small and medium-scale manufacturers be found by potential clients expediently, particularly in a digital era?"* As the digital supply chain develops to enable near-shoring and reshoring within national borders, openings in newer supply chains are created for small manufacturers to be discovered and vetted into [2]. New methods must be established to enable manufacturers, particularly small/medium business types, to access clients through automated search and discovery procedures.

Effective search and discovery in the manufacturing business can assist manufacturers in coping with the challenges presented to them amid the global pandemic. With the advent of Industry 4.0 [3], data in manufacturing systems is moving forward to be integrated into unified resource models with a closer integration of data across the product life cycle. For example, Wikidata [4] contains structured data of manufacturing certifications, services, and their corresponding relationships. In the United States, Thomasnet [5] has built a centralized database that allows small manufacturers to register on their platform with detailed data of their manufacturing services and products with built-in semantics that guides web-users based on search phrases. In sharing-economy type platforms, small manufacturers are connected to a central platform that connects clients with small manufacturers and easing a path for participant types to find each other on such platforms [6]. General purpose search engines (Google, Bing, Baidu, Yandex) also provide means for clients to find new manufacturers within geographical regions based on search keyword matching.

However, there are limitations of current Manufacturing Search Engines (MSE) [7]. First, the lack of uniform data within the search results lowers the rank of small manufacturers with very limited marketing resources, despite their strong technical capabilities. Moreover, information about a manufacturer's technical capabilities changes over time so static databases maintained by large system integrators or third-party companies are not always reliable. Acquiring such up-to-date information is crucial due to the constantly evolving nature of a resilient digital supply chain.

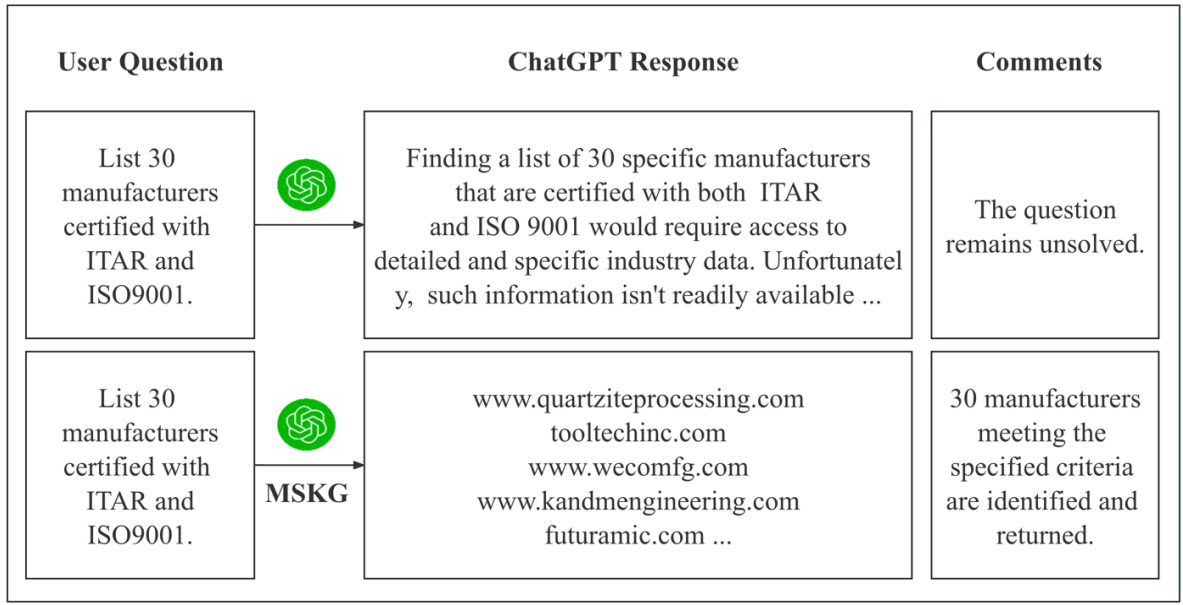

**Fig. 1.** Comparison between responses of ChatGPT and MSKG enriched ChatGPT

With the big data era, Knowledge Graph (KG) and graph-based machine learning (ML) models empower an option for extracting and training on unstructured data from vast texts, images, and videos. This representation of data in the form of KGs has been widely used in the construction of search engines as well as for powering question-answering (QA) systems. KG is built up with interconnected descriptions of entities and their interrelations to represent abstract concepts and concrete objects. A KG enables rapid growth and extension without changing the underlying schematic structure that constantly evolves and integrates data over time. KG-based search engines, such as the LinkedIn Economic Graph, Amazon Product Graph, Bing Satori, Google Vault, and Wolfram Alpha have already played an important part in helping web users find relevant results to various query terms [8]. Moreover, there is a recent trend of enabling the combination of Large Language Models (LLM) as well as KGs [9], in essence using KGs with high-quality structured data over real domains, to check the accuracy, improve the efficiency as well as retrieve more information to power LLM responses.

The advent of LLMs have made sourcing manufacturers more flexible and adjustable. By entering suitable prompts, users can interact with the model through a conversation mode, facilitating the identification of potential manufacturing partners. This dynamic approach enhances flexibility and customization, improving the efficiency and experience of the sourcing process. However, LLMs have their disadvantages which limit their use in sourcing manufacturing capabilities. For example, the question "List 30 manufacturers certified with ITAR and ISO9001." can't be addressed in GPT-4, while it can be effectively solved by combining GPT-4 with the data-rich insights from a Manufacturing Service Knowledge Graph (MSKG) (Fig. 1). The reasons that KGs can enrich LLM's responses are as follows: LLMs often lack access to domain-specific and up-to-date information found in specialized databases, unlike KGs, which are adept at sourcing varied and current data. Moreover, handling queries which demand more than general language comprehension, KGs have an advantage. They can adeptly navigate complex, industry-specific relationships by structuring entities within the graph. Additionally, LLMs, reliant on their training data, might struggle with lack of detailed contextual understanding in specialized areas. KGs, customizable with expert insights and specific data sources, provide a depth of understanding and accuracy in interpreting specialized fields beyond the capabilities of generalist LLMs. Therefore, KGs can enhance LLMs by providing up-to-date, domain-specific data and a deep understanding of complex relationships, which are essential for accurately responding to specialized queries.

This research aims to combine KGs and LLMs to facilitate the search and discovery of small manufacturing firms by their future clients. On the client side, the knowledge networks and knowledge-driven search engines will broaden access to manufacturing services enabling startups, entrepreneurs, hobbyists/amateurs, and researchers to find manufacturing services through online open resources in a more efficient way. A significant hurdle in this endeavor is the current state of manufacturing service information, which tends to be unstructured and lacks a cohesive knowledge representation that interlinks associated manufacturing resources. Our research addresses this gap by developing an MSKG that not only organizes this information more coherently but also enhances the QA capabilities of LLMs. It further supports QA functions to enrich LLMs, aiming to enhance the reliability of their responses in manufacturing service discovery.

## 2. Literature review

KGs link concepts across domains, including medicine [10], social networks [11], and chemistry [12]. KG embedding models [13] convert entities and relations into low-dimensional vectors, preserving KG structure. These models aid ML tasks like clustering [14] and link prediction [15]. Mohamed, S et al. explored Knowledge Graph Embedding for drug target prediction and clustering [16] while Wang M et al. [17] used it for medicine recommendations.

Constructing KGs from structured data is well-established but building them from unstructured data like text and multimedia is challenging due to unreliable extractions and scarce datasets. Recent efforts include COVID-KG from scientific literature [18] and an industrial KG from Chinese texts in the automobile sector [19]. Extracting accurate information from websites is tough due to noise and outdated HTML structures [20]. NLP and TLP techniques are crucial for processing large unstructured texts, employing methods like BERT [21], and GPT-4 [22]. BERT excels in various NLP tasks and has been applied in clinical information extraction [23], and mechanical equipment information extraction [24]. BART [25] is a pre-training approach generalizing BERT which achieves advanced results in the scope of dialogue, QA, and summarization tasks.

Knowledge mapping is critical in the industrial sector for visualizing knowledge, data, and relationships. Methods like LangChain [26] and LlamaIndex [27] use LLMs for data handling, complemented by ontology-driven approaches from the Industrial Ontologies Foundry [28] and manufacturing ontologies for Industry 4.0 [29]. These form the foundation of KGs for services like manufacturing service discovery [7] and equipment queries [30], aiding in industry problem-solving and decision-making [31]. Moreover, Siddharth et al. [32] have worked on extracting engineering knowledge from patents. However, literature lacks in mapping, integrating, and analyzing real-time manufacturing data. This gap stems from current LLM-based methods' limitations in industrial knowledge mapping for evolving manufacturing data integration.

The QA system combines techniques like information retrieval and knowledge-based methods to deliver precise answers. Knowledge-based QA uses KGs for sourcing answers, while graph embedding translates KG data into vectors, aiding ML, and neural networks in reasoning [33]. Knowledge-based QA can benefit LLMs since it provides a structured context that enhances the LLMs' ability to interpret and generate more accurate and contextually relevant responses.

Recent studies highlight the integration of KGs with LLMs to enhance QA systems. Daull, Xavier, et al. [34] explored the benefits of KGs in improving LLMs and reducing errors. Truong and Coleen [35] emphasized incorporating KGs for accurate response generation in LLMs. Yang, Linyao, et al. [36] proposed blending KGs with LLMs to improve response quality and factual reasoning. These advancements show promise in enhancing QA systems and accuracy, yet research on applying these methods specifically to the manufacturing industry, especially from a supply-sourcing viewpoint, remains limited. Tailoring these integrations for the manufacturing sector could significantly refine service discovery and optimize supply chain processes.

This work builds on many of the techniques used in the KG construction process and showcases an index approach, classifying and organizing manufacturing concept entities of interest contained within raw text obtained from public-facing websites of manufacturers. The contributions of our work are highlighted herein –

1. This work highlights a mechanism to extract domain specific text from several thousands of independent websites of small manufacturers, and suggests a solution to organize them, thereby enabling a more natural way of interacting with technical domain specific text for powering more natural QA types queries.
2. A method that presents a new solution to manufacturing service discovery by integrating ever-evolving KGs into LLMs, transforming the landscape of manufacturing capability identification and manufacturer recommendations.
3. A novel integration of bottom-up ontology construction with advanced machine learning models, designed to efficiently build the MSKG from both structured and unstructured data sources. This approach streamlines the assimilation of diverse data, enhancing the accuracy and relevance of the KG.
4. An advanced graph-based QA system crafted to address intricate queries related to digital supply chain networks with high reliability and interpretability. It incorporates KGs and graph embedding techniques, facilitating in-depth analysis, and enabling highly accurate similarity-based recommendations customized for the unique demands of the manufacturing sector.

The remainder of this paper is structured as follows: Section 3 provides an overview of the architecture that combines real-time MSKG and ChatGPT for manufacturing service discovery. Section 4 outlines the procedure to build a KG with content available on manufacturers' websites as well as to elucidate the use of the KG and graph-based ML framework. Section 5 gives the results of manufacturing knowledge extraction, graph embedding, and its downstream tasks as well as examples of using graph-based QA to enrich ChatGPT response. Section 6 discusses both limitations and future work of enhancing MSKG. Section 7 concludes the paper.

## 3. Architecture

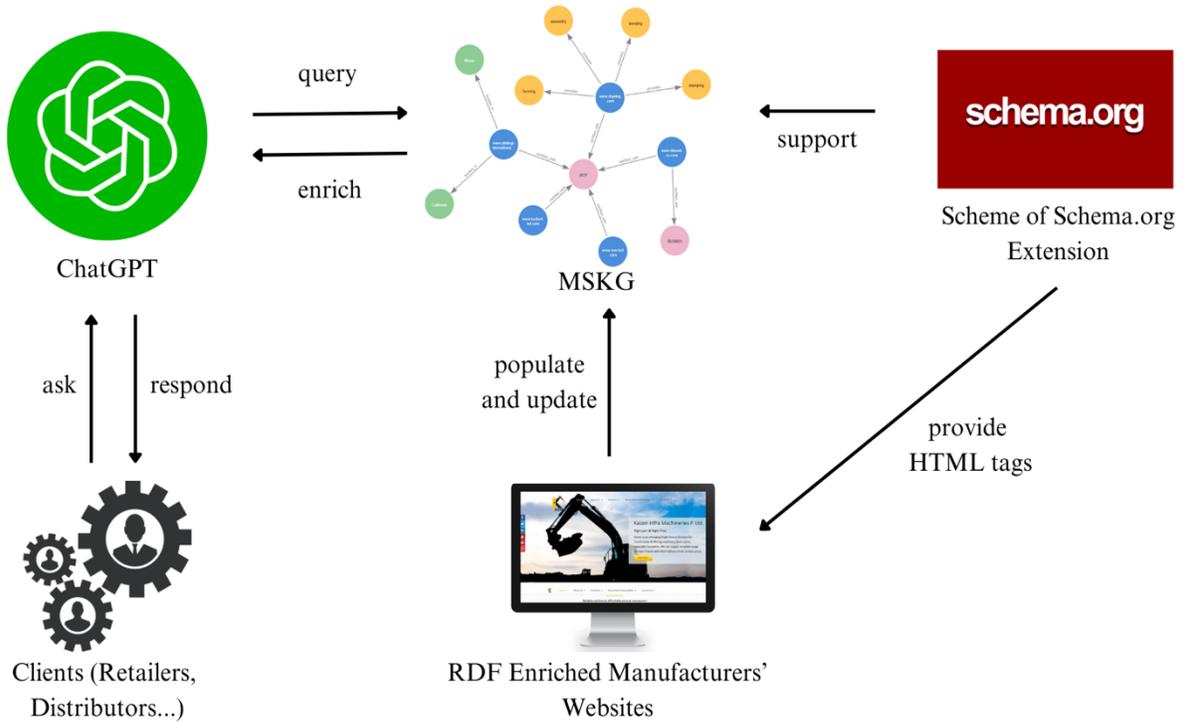

**Fig. 2.** The architecture of using MSKG to enrich ChatGPT

The comprehensive architecture integrating an MSKG with ChatGPT, designed to enhance the discovery of manufacturing services, is depicted in Fig. 2. The interaction between the client from the manufacturing industry and ChatGPT is through QA. When receiving the client's input question, the application proceeds to forward the question to the OpenAI GPT-4 endpoint, where a request is made to convert it into a query statement which can be used in the graph database. Consequently, the OpenAI endpoint responds with a query statement to retrieve the correlated manufacturing capabilities from the MSKG. The acquired data from the KG is then assisted in building a comprehensive response to address the client's initial question. In addition, MSKG can be near real-time updated by extensive manufacturers' websites [7]. The adoption of Schema.org vocabulary extension in the manufacturing domain allows manufacturers to benefit from using HTML tags to label their websites with specific manufacturing service tags. When manufacturers add these tags to their websites, the changes are reflected in MSKGs, since the extended schema relates to the ontology of MSKGs, which enables more recent and precise results for query searches.

## 4. Process Workflow

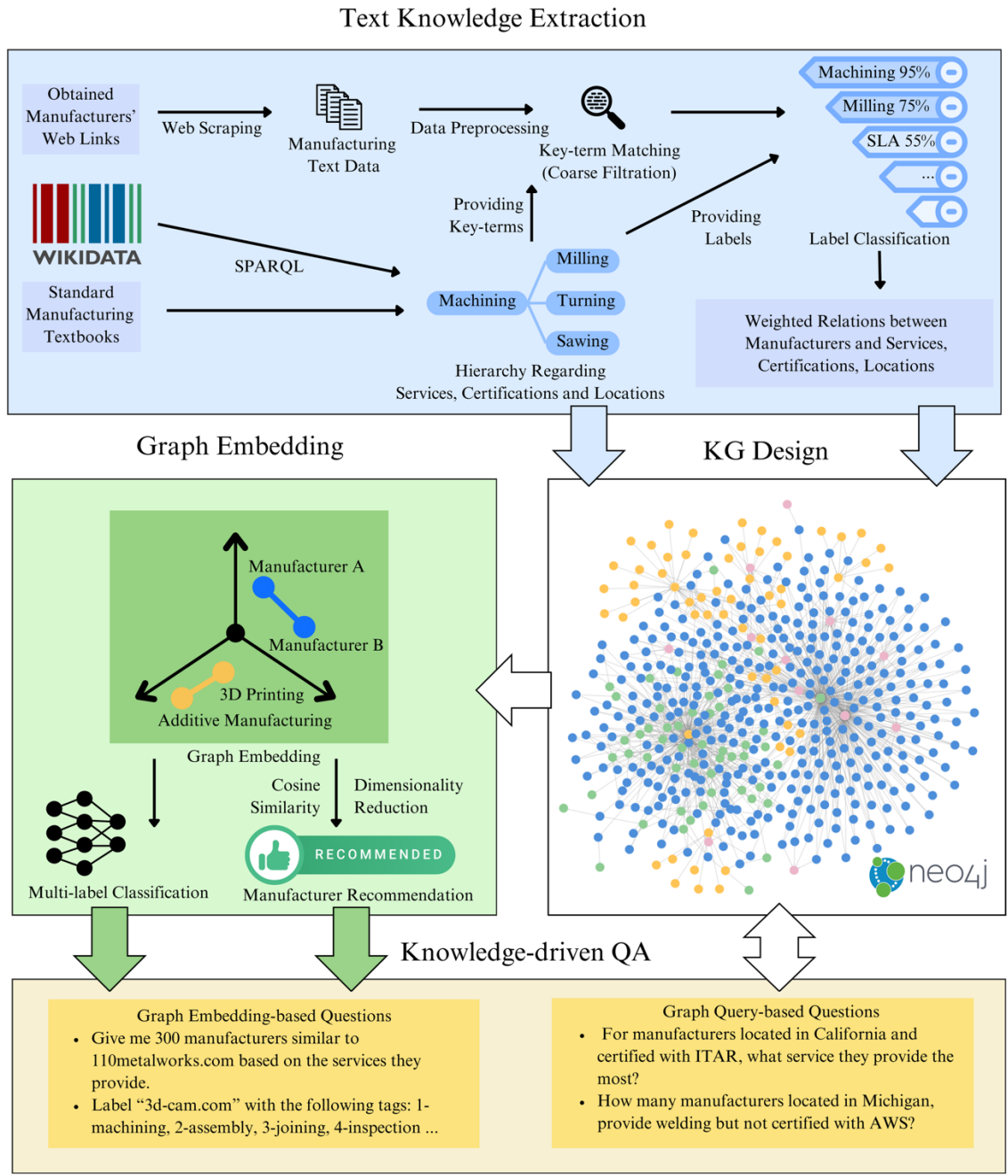

**Fig. 3.** From information on the Internet to a designed Knowledge Graph (KG) to powering a Question-Answering (QA) System.

The overall procedure to build an MSKG as well as use it to enrich QA in ChatGPT is shown in Fig. 3. The process contains four main parts, which are: Text Knowledge Extraction (Section 4.1), KG Design (Section 4.2), Graph Embedding (Section 4.3) and Knowledge-driven QA (Section 4.4). First, an information extraction procedure is conducted from manufacturers' websites and other data sources to obtain the data imported to MSKG. MSKG is built after a bulk import of entities and relations from the extraction process. This process is also guided by entities extracted from Wikidata. Based on learned graph embedding vectors from MSKG, dimensionality reduction and multi-label classification are conducted. In MSKG-based QA, not only queries from MSKG itself, but also the vectors from dimensionality reduction and trained multi-label classification model are utilized to assist ChatGPT in solving different types of questions for manufacturing capability identification. The following subsections highlight details of the process workflow in major sub-steps that were utilized in this work.

**4.1 Text Knowledge Extraction**
In the development of the MSKG, there are significant roadblocks to text extraction that we have overcome to construct the KG. One major challenge is the disambiguation of conceptually similar entities across various manufacturers' websites, where identical capabilities (services, certifications, and locations) are represented by differing text strings. Simple key term matching, while straightforward, often led to misidentifications due to its lack of contextual sensitivity. NER systems, typically optimized for well-structured texts like journal articles, prove inadequate for less organized web content and are not robust enough to differentiate manufacturing capabilities from irrelevant data. Furthermore, while contextual understanding models like BERT offer more nuanced text interpretation, the computational expense of processing extensive text with such complex models is substantial. These obstacles underscored the need for a balanced, multi-tiered approach to KG construction. An approach that combines the efficiency of key term matching for initial data filtration with the nuanced understanding of a contextual model could potentially address these challenges, offering a more precise and cost-effective solution for constructing the MSKG.

The evaluation of the MSKG employs a comprehensive set of metrics, including Accuracy, Precision, Recall, F1 score, Area Under the Receiver Operating Characteristic Curve (AUC-ROC), and Precision-Recall (AUC-PR) curve. The foundation for calculating these metrics lies in the counts of true positives (TP), true negatives (TN), false positives (FP), and false negatives (FN). These metrics offer a granular view of our model's performance, with TP, TN, FP, and FN providing direct insight into the accuracy and errors in our information extraction process. The ROC curve and PR curve further enhance our evaluation by offering insights into the model's overall predictive capabilities and the balance between precision and recall. This suite of metrics provides a thorough and quantitative assessment of our model's efficacy in reflecting real-world manufacturer data. The Accuracy, Precision, Recall, and $F_1$ score are expressed as:

$$Accuracy = \frac{TP+TN}{TP+FP+TN+FN} \quad (1)$$

$$Precision = \frac{TP}{TP+FP} \quad (2)$$

$$Recall = \frac{TP}{TP+FN} \quad (3)$$

$$F_1\ score\ = 2 \cdot \frac{Precision \cdot Recall}{Precision\ +\ Recall} \quad (4)$$

In constructing the MSKG, our approach addresses the challenges outlined in our Literature Review concerning the construction of KGs from unstructured data [19, 20]. We start with a coarse filtration process to efficiently narrow down relevant data from diverse unstructured sources. This initial step is critical in managing the volume of data and focusing on pertinent information. To further refine our process and address the complexity of entity alignment, we employ BART for label classification. This advanced NLP technique rectifies potential misidentifications from key term matching and adds a layer of semantic understanding. The combination of coarse filtration with BART's semantic analysis ensures a robust and nuanced construction of the MSKG, aligning with the complexities of building KGs from varied, unstructured data sources.

To populate the MSKG, text data from the front pages of manufacturers' websites (Total = 17,230) in the US and Canada, Wikidata, and standard manufacturing textbooks are extracted using the predefined ontology that categorizes entities into Manufacturer Name, Service, Certification, and Location (shown in Table 1). The service, certification, and location information, extracted from Wikidata and standard manufacturing textbooks, are not only integrated into the MSKG as entities and relationships, but also used as pivotal key terms for the coarse filtration process and as extraction labels in the label classification phase. Weighted relationships linking Manufacturer Names to the other entities are imported to the graph after label classification, which effectively reflects the interconnectedness of the data.

**Table 1** Extracted entity types

| Entity Type | Description | Example |
|---|---|---|
| Manufacturer | Web URL of Manufacturers | acufab.com, 3axis.us, 1stmanufacturing.com |
| Service | Manufacturing Service Names selected from Wikidata | milling, heat treatment, assembly |
| Certification | Manufacturing Certifications selected from Wikidata | ISO9001, AS9100, CMMI |
| Location | All the States in the U.S and Canada | North Carolina, British Columbia, Ohio |

The coarse filtration involves implementing key term matching to isolate essential N-grams (N<10) from the collected text gathered through the web scraping process, focusing on terms directly related to the three entity types: Service, Certification, and Location. For instance, terms like "ISO 9001" or "heat treating" are used as anchors in this stage. Coarse filtration effectively reduces the volume of data by filtering out non-essential information, setting the stage for the more nuanced label classification process. Furthermore, this initial stage of filtration lays the groundwork for establishing certain key relationships within the MSKG. The assigned weight of 0.8 for the relationships serves as an empirical confidence score, not as a true probability or a model-derived predictive probability. This heuristic is chosen based on its demonstrated reliability in indicating accurate service relationships, particularly for services prominently listed on a manufacturer's homepage- observed during the validation phase. The weight's effectiveness is further supported by its ability to outperform label classification in predictive accuracy and its correspondence with the highest score for incorrect classifications, from the location and certification extractions validation. This makes the weight of 0.8 not only a robust indicator of service provision by manufacturers but also a conservative benchmark for maintaining the integrity of data within the KG.

Label classification utilizes a BART-large-mnli model, equipped with a zero-shot text classification pipeline. In this process, text embedding vectors derived from the data filtered through key term matching, are obtained using BART and then input into a multi-label classification model. This model is capable of classifying text into predefined labels without requiring any prior data. The effectiveness and reliability of employing BART for this purpose are evidenced in our validation results, with a representative sample of the outcomes presented in Table 2.

By combining coarse filtration with label classification, our method goes beyond mere key-term matching, enabling the extraction of a broader range of relationships from text. This is due to its ability to contextually understand different descriptions and variations of the same term. For example, if "ISO 9001" is selected as a key term, only N-grams containing the content of "ISO 9001" can be tagged. However, by using label classification, "ISO9001", "ISO_9001" and "iso 9001" are extracted as well. Another example is an N-gram including "heat treating" which can be recognized as a similar expression of "heat treatment" by label classification. Additionally, label classification is used to make up for the misidentification of key terms matching with semantic understanding. An example of the service extraction result is shown in Table 2. The abbreviations of certifications and locations selected as key terms of coarse filtration, such as "CMMI" and "CA", may match irrelevant N-grams, while the classification can filter out the N-grams that are less likely to correlate the manufacturers' data.

**Table 2** A service extraction sample

| "Manufacturer" | Weight | Extraction-Label |
|---|---|---|
| 3dmouldmfgltd.com | 0.46 | additive manufacturing |
| 3dmouldmfgltd.com | 0.80 | forming |
| 3dmouldmfgltd.com | 0.80 | machining |

In the experimental setup, we analyze a 1% sample from 17,230 manufacturers' websites, focusing on extracting service, certification, and location information. Each extracted item is manually classified as either positive or negative, forming a binary classification. The weights are assigned as probability estimates: a default of 0.8 for relationships identified during coarse filtration and 0.2 for instances where no relationship was found. These outcomes allow us to evaluate the effectiveness of our extraction process, providing insights into the accuracy and reliability of identifying relevant information from the manufacturers' websites.

**4.2 KG Design**

The KG [37] involves 4 kinds of node labels and 4 types of relation labels. An example with all types of nodes and relationships is shown in Fig. 4. Each blue node labeled "Manufacturer" stands for a manufacturer or supplier business entity. An orange "Service" node means a type of manufacturing service or capability that can be provided by "Manufacturer". A node of "Certification" in pink means the types of certifications or industry standards that a "Manufacturer" is certified with or registered to have. The green node named "Location" is a state in America or Canada in which the "Manufacturer" is located. Three different relation types that connect between "Manufacturer" and the other three types of nodes are 1) "provides"; 2) "certified_with", and 3) "located_in". Besides this, another relation "subclass_of" is used to describe the relationship between two "Service" types, meaning when one manufacturing service belongs to another general manufacturing service type.

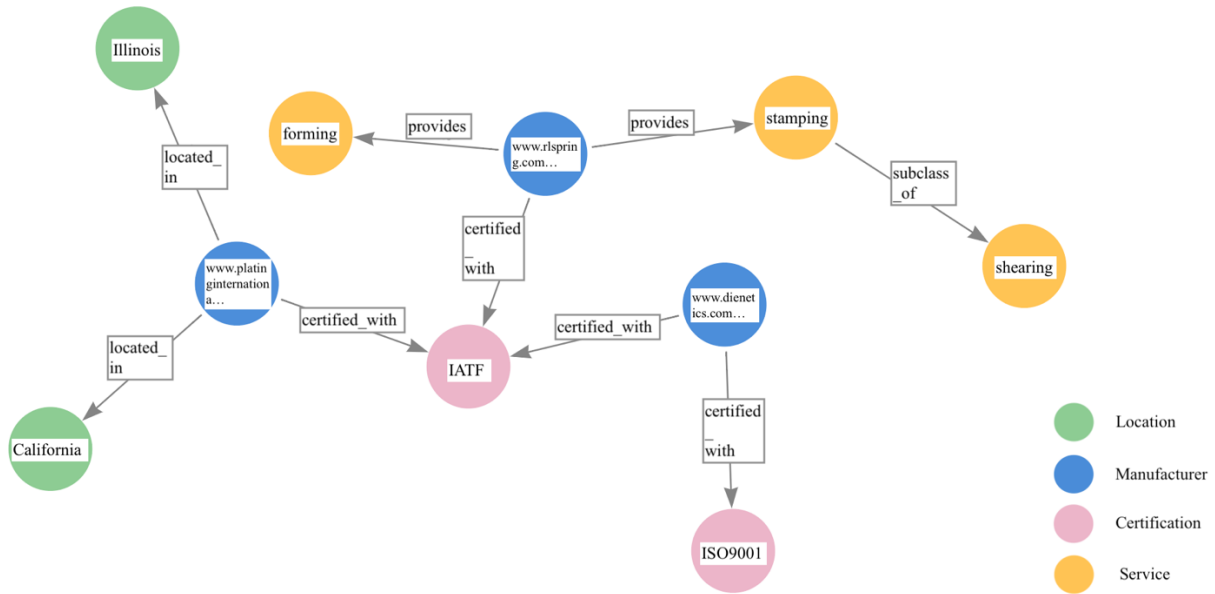

**Fig. 4.** General MSKG structure

Each "Service" entity has a unique ID, which is similar to the ID provided for the same service term on Wikidata. Each Wikidata entry term has a unique ID associated with it. Referring to the same Wikidata ID in our MSKG, it enables our MSKG to be connected within the more general KG built by Wikidata. Each relation has an attribute of weight. The total number of entities and relationships is presented in Table 3. The KG is designed to serve multi-conditioned queries and secondary relation queries. Examples of multi-conditioned queries are "Give a list of 3D-printing and 5-axis-milling manufacturers." and "What manufacturer can provide both injection-molding and sand-casting services?". Examples of secondary relation queries are: "What is the geolocation distribution of suppliers who provide sand-casting services?" and "Give the list of all the certifications obtained by suppliers of additive-manufacturing services". These questions can be answered by matching conditional sub-graphs of the KG and returning specified entities or relationships in the subgraphs via Neo4j query language (Cypher).

**Table 3** Total entities and relationships of KG

| Entity Type | Number of Entities | Relationships | Total Relationships |
| --- | --- | --- | --- |
| Manufacturer | 13,085 | provides: Manufacturer → Service | 39,761 |
| Service | 77 | subclass_of: Service → Service | 76 |
| Certification | 15 | certified_with: Manufacturer → Certification | 3968 |
| Location | 63 | located_in: Manufacturer → Location | 14,806 |
| Total | 13,240 | | 58,521 |

## 4.3 Graph Embedding

To analyze the similarity and diversity among the manufacturers based on the manufacturing services provided, labeled entities of "Manufacturer" and "Service" and their correlated weighted relationships were extracted as a sub-graph from the larger KG for graph embedding. Two graph embedding methods are applied in this paper, which are Node2Vec [38] and GraphSAGE [39]. They are used to learn embedding vectors from a heterogeneous subgraph of the KG considering all the classification weights obtained from the text knowledge extraction process. The dimensions of the learned embedding vectors are set to 100. Two graph embedding downstream tasks are conducted, which are manufacturer recommendation and multi-label classification.

Node2Vec and GraphSAGE are selected due to their distinct strengths in graph embedding. Node2Vec is adept at capturing intricate connectivity patterns through its neighborhood sampling technique, ideal for representing complex relationships between manufacturers and their services. GraphSAGE, with its inductive learning and neighbor sampling approach, excels in handling heterogeneous graphs like ours, generating scalable embeddings. By applying these methods side by side, we aim to compare their performance and effectiveness in accurately representing our dataset, thus enabling a thorough evaluation of their respective capabilities in the context of our analysis.

In supply chain management, selecting suitable manufacturers is crucial for achieving operational efficiency and customer satisfaction. For manufacturer recommendation, dimensionality reduction is used to visualize the clusters of manufacturer nodes in MSKG, which shows service-related characteristics of the manufacturers directly as well as help with the discovery of similar manufacturers. T-Distributed Stochastic Neighboring Entities (T-SNE) is chosen for reducing the dimensions for visualization [40] in a 2D space. The vectors generated from dimensionality reduction are used for similarity-based manufacturer recommendations. Cosine similarity is used to calculate the similarity between manufacturer nodes.

Multi-label classification [41] is a variant of the classification problem where multiple nonexclusive labels may be assigned to each instance. In this work's context, it is used to gain a rapid and preliminary understanding of manufacturers' capabilities. The trained classification model can predict and return the probabilities of each manufacturer correlated with a finite set of categories (N=10), which are determined by the hierarchy of services in the KG. The 10 categories are 1-machining, 2-assembly, 3-joining, 4-inspection, 5-forming, 6-molding, 7-casting, 8-additive manufacturing, 9-heat treatment, 10-other. When queried about "3d-cam.com", the system categorizes its services into four key areas: "machining" for precision-crafted parts, "molding" for high-volume production, "casting" for complex shapes, and "additive manufacturing" for 3D model-based creation. This classification succinctly highlights the company's diverse and specialized manufacturing capabilities.

A Multi-Layer Perceptron (MLP) model, specifically designed for multi-label classification tasks, is trained to estimate the likelihood that various manufacturers are associated with each of the pre-defined service categories. The learned graph embedding vectors from Node2Vec are randomly separated in a ratio of 80:10:10 for datasets of training, testing, and validation. The MLP model comprises 100 input neurons, a hidden layer of 20 neurons leveraging Rectified Linear Units (ReLU) activation function [42] and 'he_uniform' initialization, and an output layer of 10 neurons with sigmoid functions [43] for binary classification. Optimization is carried out using the Adam algorithm [44] coupled with binary cross-entropy loss [45], capitalizing on their collective strengths in gradient handling and probabilistic output calibration. The model's robustness is assured through 100 training epochs, a batch size of 32 for gradient estimation, and a stringent 10-fold cross-validation protocol reiterated thrice, ensuring a comprehensive evaluation against overfitting, and confirming the model's predictive consistency.

### 4.4 Knowledge-driven QA

Constructing QA systems for manufacturing service discovery presents unique challenges, primarily due to the complex and dynamic nature of the manufacturing industry. Key issues include integrating detailed, industry-specific data into KGs and ensuring they are continuously updated to reflect new advancements and market trends. Furthermore, accurately modeling the intricate relationships within the manufacturing supply chain in a KG is complex. The high accuracy and reliability demand in manufacturing, coupled with limited access to proprietary data, add to the difficulty in creating effective QA systems for this sector.

There are several ways to evaluate a QA system, including Mean Reciprocal Rank (MRR) [48], Precision at N, Recall, $F_1$ score, and human evaluation. Precision at N measures the proportion of the Top N answers returned by a recommender system correctly. To evaluate the performance of our similarity-based manufacturer recommendation, we use Precision at N metric (N=10,100,300), which serves as an indicator of the system's ability to return relevant answers in the list of generated results. Precision at N (P@N) is expressed as:

$$P@N = \frac{N_{relevant}}{N_{total}} \quad (5)$$

$N_{relevant}$ is the number of services relevant to a target manufacturer among the Top N results. $N_{total}$ is the number of services provided by Top N results. Relevant services can be the services from the target manufacturer or the subclass services of the target's services. The reason for calculating P@N in this manner arises from the fact that the manufacturer recommendations are formulated not solely based on exact service matching, but also by considering approximate search matches, as well as interrelationships among the services.

Besides, MRR is also used for evaluating the effectiveness of manufacturer recommendations. It is especially for scenarios where the recommendation system can suggest a list of manufacturers' websites, and the user is expected to interact with only one of them. MRR is expressed as:

$$MRR = \frac{1}{|Q|} \sum_{i=1}^{|Q|} \frac{1}{rank_i} \quad (6)$$

MRR is the average of the reciprocal ranks of results for a set of queries Q, where $rank_i$ refers to the rank position of the first relevant manufacturer in Top N results for the i-th query.

Choosing P@N and MRR for evaluating manufacturer recommendations in the QA system aligns well with the need for precise and top-ranked responses in manufacturer discovery. P@N assesses the accuracy of the top recommendations, crucial for users seeking specific manufacturers, while MRR evaluates how effectively the system identifies the most relevant manufacturer as the first choice. For other queries, a manual comparison of responses from the MSKG against those from LLMs allows for a detailed assessment of the system's capabilities in handling a variety of queries, ensuring a comprehensive evaluation of its overall performance in the manufacturing context.

The research discussed in the Literature Review, encapsulated in studies [34-36], emphasizes enhancing LLMs with KGs for more precise and dependable responses. Yet, these developments have not been directly adapted to the manufacturing services sector, particularly in the context of supply-demand sourcing. This presents a unique opportunity to customize these KG-LLM integration strategies for the distinct requirements of manufacturing service discovery, aiming to boost both accuracy and efficiency in this area.

Diverging from the typical KG-based QA systems, the QA component introduced in this study is crafted to cater to queries about sourcing manufacturers and their services, addressing a specific need in the manufacturing industry. We develop two tiers of questions: one based on graph queries (easy-level) and the other on graph embeddings (hard-level). These are used to evaluate the responses from ChatGPT alone versus ChatGPT combined with MSKG, aiming to gauge the effectiveness of our QA system in addressing these benchmark inquiries [46].

When receiving graph query-based questions, by using Langchain with a customized prompt, users' queries are identified as easy questions and then transferred into cypher queries. Langchain library connects ChatGPT, with a Neo4j graph database. The schema of the Neo4j database is refreshed to ensure data consistency. The cypher queries are utilized to search the entities and relationships in the MSKG, which are sent back to ChatGPT to help with building reliable and interpretable answers. Fig.5. is an example of solving the "How many manufacturers provide additive manufacturing in each state?" with MSKG-enriched ChatGPT.

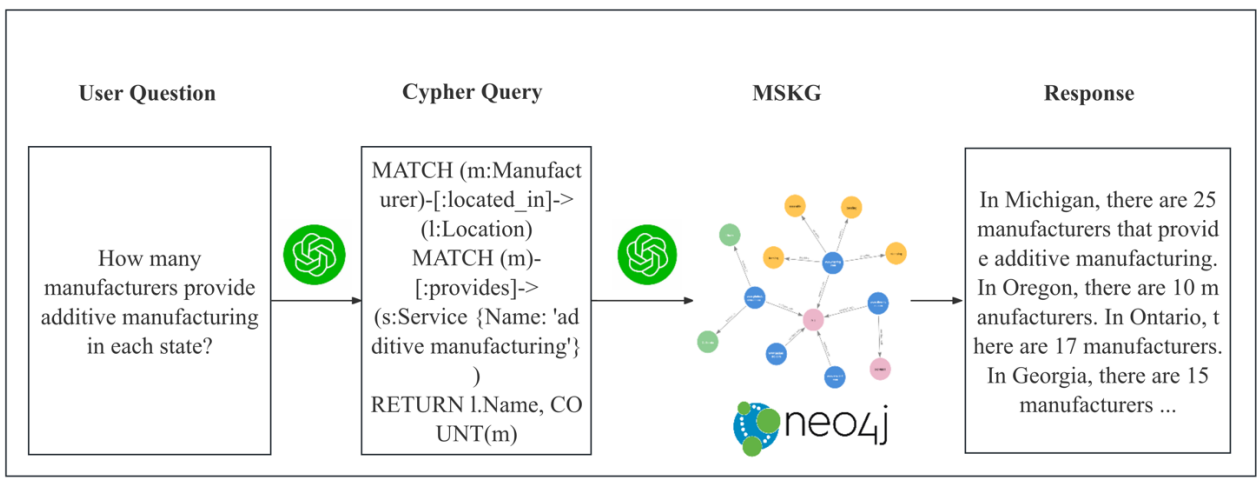

**Fig. 5.** Solving an easy-level question combining MSKG and ChatGPT

The graph embedding-based questions are those when querying a single KG may not directly answer them. They are identified as 'hard questions' by Langchain with a prompt. Our graph embedding downstream tasks, as mentioned in the manufacturer recommendation and multi-label classification, are applied to process these questions. For instance, when the query "Give me 300 manufacturers similar to 110metalworks.com based on the services they provide." is proposed (shown in Fig.6.), the system identifies it as a 'hard question', then '110metalworks.com' is extracted as the name of a manufacturer node. A cypher query is then executed to retrieve the ID of the corresponding entity from MSKG, ensuring an exact match with a node named '110metalworks.com' that already has a pre-calculated embedding vector. The similarities of graph embedding vectors between the node, and all the other manufacturer nodes are computed, sorted as well as returned with Top 300 results to support the response. Another example is to label a list of manufacturers with predefined tags. The trained multi-label classification model can predict the labels and return the results to the chatbots.

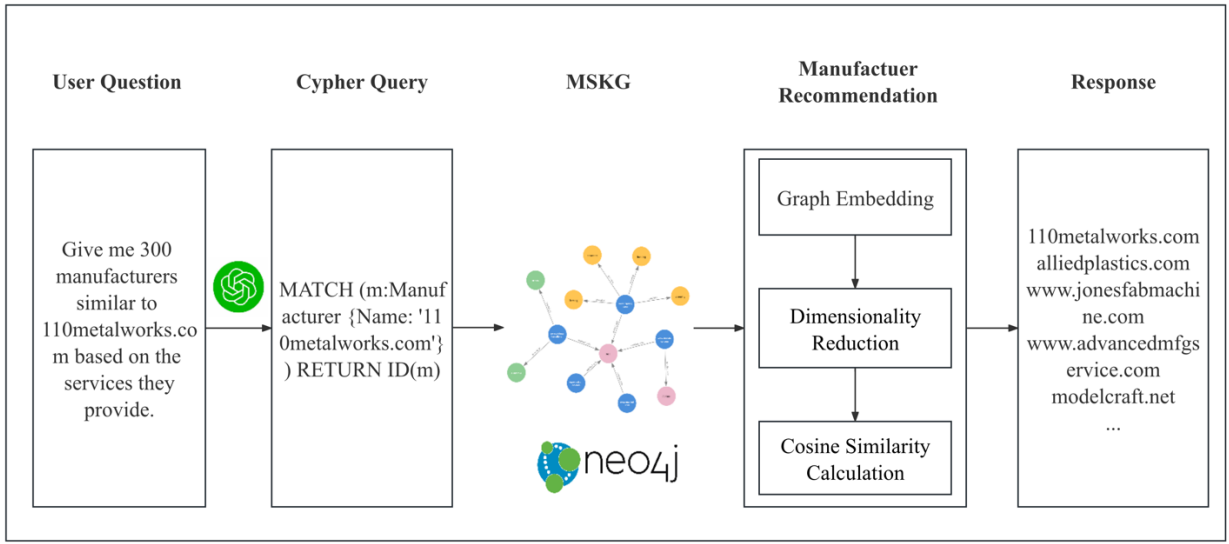

**Fig.6.** Solving a hard-level question combining MSKG and ChatGPT

In the experimental setup, we evaluate the graph-based QA system against LLMs by comparing responses to 15 key questions in manufacturing service discovery. For manufacturer recommendation-related questions (Q13 and Q14), we assess the answers using Precision at N (P@N) metrics for N=10, 100, 300, alongside MRR. By comparing the MSKG-based QA system with standard LLMs, our methodology aims to highlight the potential benefits and improvements in accuracy and efficiency that can be achieved through this specialized integration for the manufacturing sector.

## 5. Results

### 5.1 Result of Extraction Validation of Text from Websites of Manufacturers

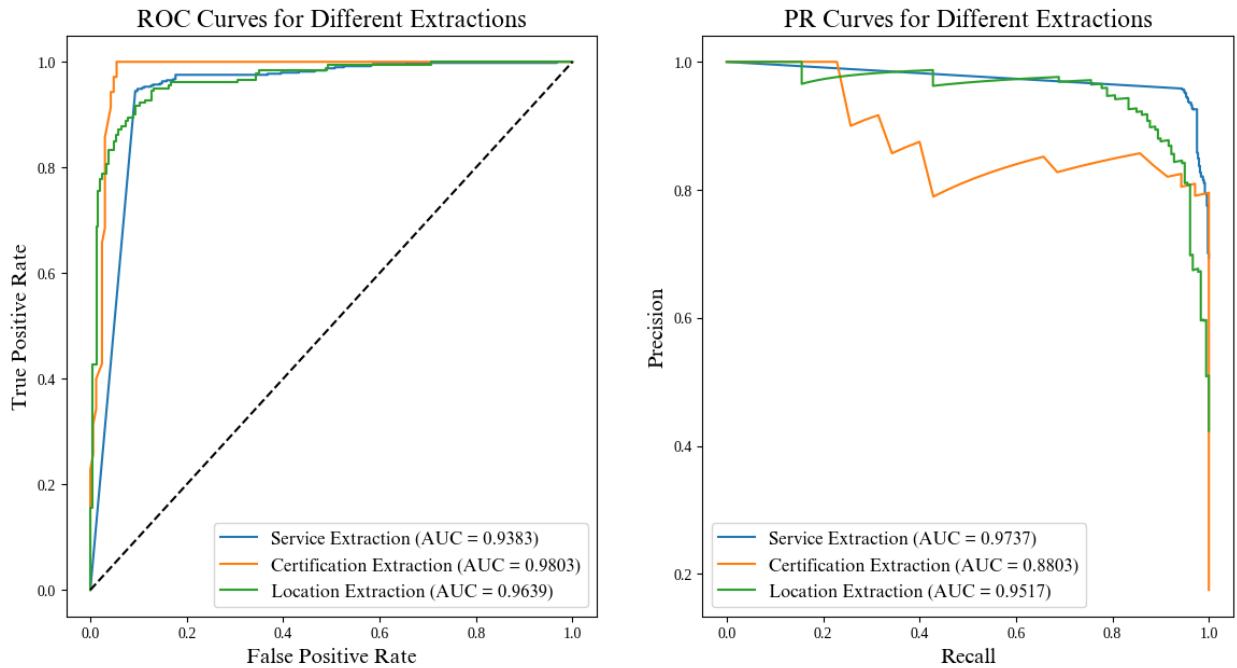

**Fig.7.** ROC curves and PR curves of text knowledge extraction

In our extraction results, the number of negative classes may be much bigger than positive classes due to the incompleteness and diversity of text information on the main page of manufacturers' websites. PR curves are necessary to be computed to show the reliability of our model with ROC curves. The ROC curves as well as PR curves with AUCs of 1% extraction results are shown in Fig.7. Thus, all the extraction models have good performance which makes directly applying a trained model applicable. The certification extraction model, highlighted in orange, shows the best performance with the highest AUC-ROC score, while the location extraction model is the least effective. The performance variation likely results from the different entity frequencies in the text, also affecting the certification model's AUC-PR.

The selection of cutoff values for the precision, recall, and $F_1$ score calculations are carefully optimized for maximum accuracy across different types of data extraction. We have identified the ideal cutoff points for service, certification, and location extraction at 0.4, 0.25, and 0.4, respectively. Implementing these specific thresholds for each extraction model, our approach achieves outstanding overall data extraction performance, boasting an AUC-ROC of 95.91%, an AUC-PR of 95.26%, an accuracy of 89.10%, an $F_1$ score of 90.76%, a recall of 89.10%, and a precision of 93.92%. Table 4 shows a comparison of our data extraction results to form MSKG when compared to other KG construction efforts. The existing KG data extraction methods are not used for building the MSKG because they are designed for structured texts and perform poorly on the unstructured data of company websites, failing to precisely identify manufacturing capabilities from other recognized entities. Our KG data extraction model has better performance since it contains both coarse filtration by key term matching and text classification. Some of the false-positive entities recognized by coarse filtration can be filtered out from the text label classification, which increases the ability of the model to capture as many positive classes as possible. The lists of key-term matching, and labels of text classification are highly correlated, which motivates our model to find fewer entities precisely than finding incorrect entities.

**Table 4** Comparison of evaluation metrics of KG data extraction

| Model | Precision | Recall | $F_1$ score |
|---|---|---|---|
| PSL[48] | 78% | 94% | 85% |
| DeNERT-KG[49] | 72 % | 73% | 72% |
| Automobile Industry Data Extraction[50] | 83% | N/A | 78% |
| FabNER [51] | 93% | 84% | 88% |
| Our Work | 94% | 88% | 91% |

### 5.2 Results of Graph embedding and Its Downstream Tasks

A 100-dimensional vector space for all entities labeled "Manufacturer" in the graph is obtained from the Node2Vec and GraphSAGE embedding results. The dimensionality of the embedding vectors is chosen to be 100 for balancing the trade-off between computational resources and performance based on the observation in [38]. T-SNE is used to project the high-dimensional vector to a 2D Space. Fig. 8. and Fig. 9. compare the efficacy of Node2Vec and GraphSAGE in distinguishing service-related attributes of manufacturers. Utilizing the Elbow Method [53], we determine that the optimal number of clusters is 4. We assessed clustering performance using Silhouette scores [54], which gauge the proximity of points within a cluster to those in adjacent clusters. With Silhouette scores of 0.6949 for GraphSAGE and 0.4059 for Node2Vec, we observe a markedly better clustering definition for GraphSAGE. This suggests that GraphSAGE provides a more distinct and coherent grouping of service characteristics, underscoring its potential for more nuanced differentiation of manufacturers' services in our analysis.

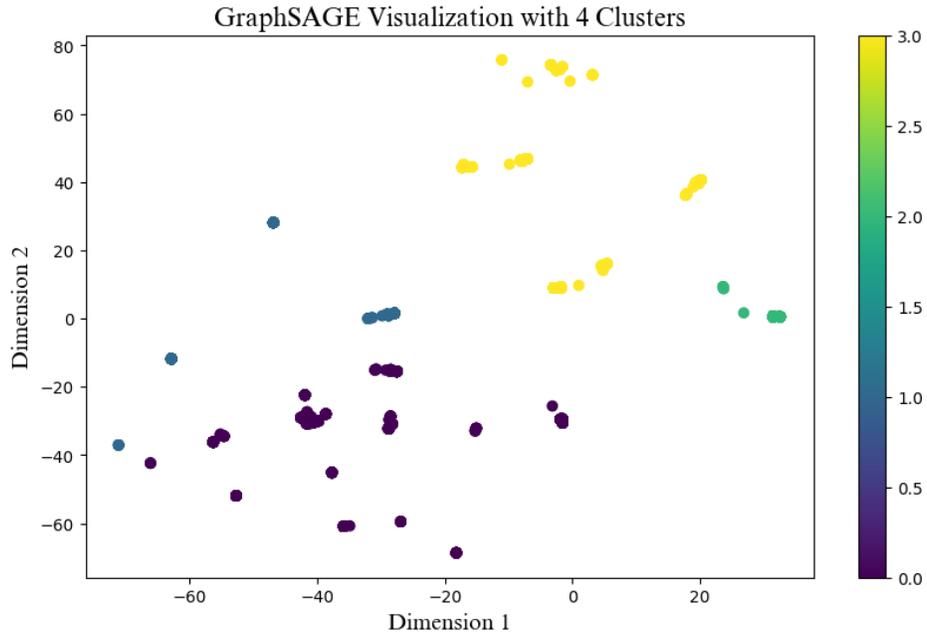

**Fig. 8.** T-SNE visualization using GraphSAGE for service-related attributes of manufacturers

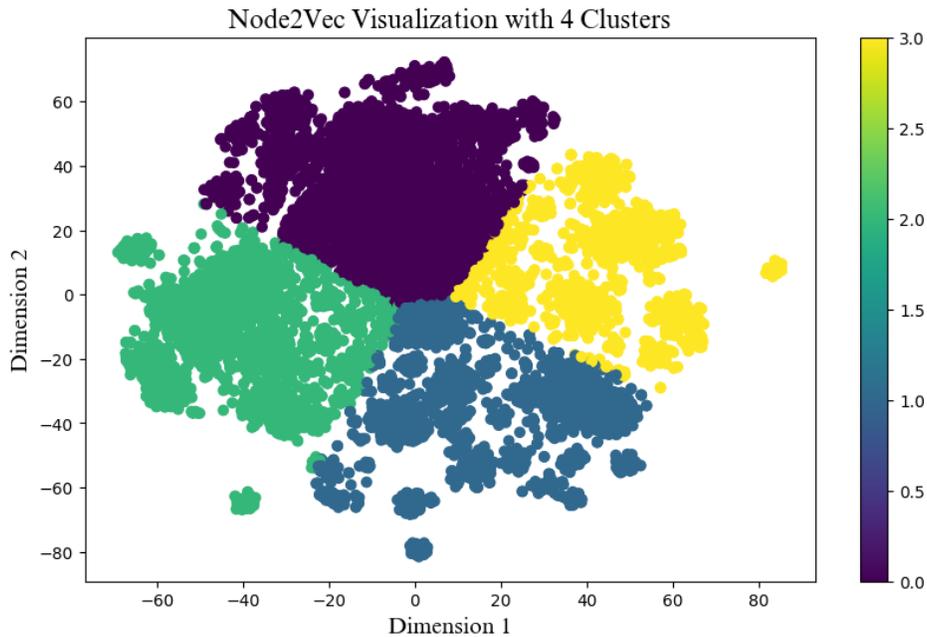

**Fig. 9.** T-SNE visualization using node2Vec for service-related attributes of manufacturers

For the task of multi-label classification, Node2Vec-generated embedding vectors are employed. These vectors are then trained and assessed through an MLP model, utilizing a repeated k-fold cross-validation approach for robust evaluation [55]. The training accuracy is 98.90%. The accuracy, $F_1$ score, recall, and precision of predicting multi-labels of validation datasets are 98.72%, 94.62%, 99.93%, and 89.85%, respectively, which shows good performance of multi-label predictions.

## 5.3 Evaluation of QA based on MSKG

In the Appendix, we include a detailed comparison of typical questions for manufacturing service discovery along with their corresponding answers derived from the MSKG, benchmarked against responses from GPT-4 (with browsing and analysis). The evidence supporting all the answers is enclosed in the supplementary materials [46]. According to the Appendix, for the simplest questions, either GPT-4 or MSKG can provide answers. For Q1-Q15, the integration of GPT-4 with insights from the MSKG is imperative to tackle queries that go beyond GPT-4's standalone capabilities. The reasons for this necessity can be outlined as follows: Firstly, MSKG answers draw from a foundation of structured data, providing a higher level of reliability and interpretability. In contrast, ChatGPT's responses often rely on pattern recognition and estimations, leading to qualifiers like "general," "likely," and "might," indicating a lack of definitive reliability. Secondly, MSKG empowers the utilization of intricate queries within the graph, enabling the resolution of questions that surpass GPT-4's capabilities. This functionality ensures that even the most complex aspects of inquiries can be effectively addressed.

According to the evaluation results of manufacturer recommendations in Table 5, The results show that for Q13, GraphSAGE slightly outperforms Node2Vec at P@10, but both methods perform similarly at P@100 and P@300, and both achieve a perfect MRR. For Q14, Node2Vec is superior, particularly at P@10, and achieves a perfect MRR, while GraphSAGE's MRR drops to 50%. The distinction between Q13 and Q14 pertains to the quantity of services offered by the target manufacturer, which are 3 and 6, respectively. This difference proves that our manufacturer recommendation function exhibits better performance when the target manufacturers provide a smaller number of services.

**Table 5** Evaluation of manufacturer recommendations

| Evaluation Metrics | Q13 with Node2Vec | Q13 with GraphSAGE | Q14 with Node2Vec | Q14 with GraphSAGE |
|---|---|---|---|---|
| P@10 | 80% | 90% | 70% | 50% |
| P@100 | 83% | 82% | 71% | 73% |
| P@300 | 84% | 81% | 79% | 78% |
| MRR | 100% | 100% | 100% | 50% |

## 6. Discussion

This work employs a bottom-up approach to gather raw data from manufacturers' websites to build a knowledge graph (KG) with four entity types and their corresponding relationships. However, challenges arise due to the rudimentary website designs, leading to the extraction of front-page information from only 13,000+ out of approximately 17,000 companies, as some sites use animated gifs, videos, or images as their front-page content. Additionally, services and certifications presented as images on websites complicate text extraction, hindering search engine indexing algorithms. The lack of basic Search Engine Optimization (SEO) code further prevents accurate classification by search engines. To address this, developing and widely adopting manufacturing-specific schema.org extensions within manufacturers' website content can enhance the indexing and linking of manufacturing capabilities both nationally and globally.

Our analysis used Node2Vec and GraphSAGE on an undirected homogeneous graph, overlooking the MSKGs' directional relationships and labels. Our MSKG, a directed heterogeneous graph, holds crucial asymmetric semantic details in its entity relationships, suggesting areas for future research. Future work should aim to improve tasks like manufacturer recommendation and multi-label classification by developing embeddings for our specific graph type. Proposed solutions include utilizing frameworks like DGGAN [56], DGE [57], and HetGNN [58] for advanced graph representations. Additionally, the lack of customer activity data in the MSKG, crucial for accurate manufacturer recommendations, presents a cold start problem. This challenge could be addressed by integrating side information into the MSKG's embeddings [59].

To boost MSKG's potential, it's crucial to improve web link capturing and refine image classification on manufacturers' websites using deep learning techniques, such as CNNs [60]. This enhances the MSKG with valuable technical imagery and broadens its scope to include manufacturers' sectors and materials, establishing their interconnections. Integrating real-time, diverse data into KGs presents challenges, especially in cross-domain knowledge integration [61]. Future efforts should focus on improving LLMs' understanding of KG contexts, potentially through targeted fine-tuning and pretraining strategies that leverage graph information, enhancing their interpretive and reasoning skills regarding KGs [62].

## 7. Conclusion

This paper introduces a framework that leverages near real-time updated KGs to potentially enrich LLMs in manufacturing service identification and manufacturer recommendation. This work lays out a bottom-up approach to build MSKG from both unstructured data and structured data, using existing ontologies from sources like Wikidata. The accuracy and efficiency of data extraction for the KG have been significantly improved, thereby enhancing the overall quality and reliability of the MSKG. MSKG is built specifically for manufacturing services containing four entity types and their corresponding relationship types, with a total of 13,240 entities and 58,521 relationships from text content available from a subset of manufacturers available in North America. We contribute a publicly available dataset containing manufacturers' weblinks, manufacturing services, certifications, and locations for further study. The evaluation results of the data extracted and integrated into the KG reflect that the constructed KG has high reliability.

The KG and the learned graph embedding vectors are used to support QA in ChatGPT. Leveraging the transformation between human natural language and graph query language, the knowledge can be retrieved from MSKG to answer the questions from ChatGPT's clients in the manufacturing industry. Based on graph embedding, manufacturer recommendation, and multi-label classification results summarize the commonalities and features of the manufacturer which can greatly support ChatGPT in the identification and recommendation of manufacturers. The comparison results between LLMs and the proposed MSKG-based QA indicate the latter's capability to effectively respond to complex queries in support of sourcing for small-scale manufacturing businesses.

The scale of MSKG can be further expanded to include more entity types and relationships that can span domains that are adjacent to a manufacturing supply chain (for example, the business and legal domain) or specific industry supply chains such as those serving microelectronics, aerospace and medical sector. It can also be expanded to include more fine-grained details within the MSKG to power richer searches. Any future framework will be an integration of knowledge models with LLMs. Knowledge models serving different sections in the manufacturing industry can be constructed through automated mechanisms by fine-tuned LLMs.

**Acknowledgement**
The authors would like to thank the anonymous reviewers for their invaluable comments that led to an improved article. We gratefully acknowledge support from NSF Grant# 1937043 and partial support from NSF Grant#2032043D.

**Data availability**
The MSKG containing 13240 entities and 58521 relationships is available [37]. Examples of QA in MSKG are available [46].

**Declaration of Generative AI and AI-assisted technologies in the writing process**
During the preparation of this work the author(s) used ChatGPT in order to improve readability and language. After using this tool/service, the author(s) reviewed and edited the content as needed and take(s) full responsibility for the content described in the article.

**Appendix**
As mentioned in Section 5.3, Table A.1 provides a comprehensive comparison and detailed analysis of the questions and answers pertinent to the manufacturing industry. This table contrasts the responses generated by MSKG with those from GPT-4, aligning both sets of answers against the original data sourced directly from manufacturers' websites. This comparison highlights the superior accuracy and relevance of the MSKG-driven responses, clearly demonstrating its enhanced efficacy in addressing industry-specific queries when compared to GPT-4's outputs.

**Table A.1** Examples of QA in MSKG

| Level | Typical Questions | Answers using MSKG | Answer from GPT-4 with Browsing and Analysis | Comments |
|---|---|---|---|---|
| N/A | What are the locations related to "1stmanufacturing.com"? | California, South Dakota | Rocklin, California: First Manufacturing began its operations part-time in January 2003 in Rocklin, California, with basic manufacturing equipment like a manual mill and a manual lathe. Sioux Falls, South Dakota: In October 2003, the company moved to Sioux Falls, South Dakota, where it started to expand its machining capabilities. ... | The questions can be answered by either MSKG or GPT-4. |
| | Is "1stmanufacturing.com" certified for AS9100? | No | The search results did not provide specific information about whether "1stmanufacturing.com" or First Manufacturing LLC is certified for AS9100... It may be beneficial to contact the company directly or consult industry-specific directories for this information. | |

| | Question | MSKG-GPT-4 | GPT-4 | Remarks |
|---|---|---|---|---|
| Easy (Graph Query-based) | 1. List 50 manufacturers certified with ITAR. | janewaymachine.com qtmfg.com www.klsteven.com ... | Listing 50 specific manufacturers certified for the International Traffic in Arms Regulations (ITAR) is a challenging task, given that ITAR certification information is not typically publicly disclosed in a comprehensive list. However, I can guide you on how to find manufacturers that are ITAR certified:1. **Industry Directories** 2. **ITAR Registered Companies**... | The questions exceed GPT-4's capabilities alone but can be effectively answered by combining GPT-4 with the data-rich insights from the MSKG. |
| | 2. List 30 manufacturers certified with ITAR and ISO9001. | www.quartziteprocessing.com tooltechinc.com www.wecomfg.com ... | Finding a list of 30 specific manufacturers that are certified with both ITAR and ISO 9001 would require access to detailed and specific industry data. Unfortunately, such information isn't readily available in an open-source format due to the sensitive nature of ITAR-related information and the proprietary nature of business certifications. To obtain this information, consider the following avenues:1. **Industry Trade Associations**2. **Certification Bodies**… | |
| | 3. List 50 manufacturers which provide welding as well as certified with American Welding Society (AWS). | thieman.com www.southernmetalfab.com carolinafab.com ... | Finding a list of 50 specific manufacturers that provide welding services and are certified by the American Welding Society (AWS) would typically require access to industry-specific databases or membership directories. However, I can offer guidance on how to locate such manufacturers:1. **AWS Corporate Members Directory** 2. **AWS Certified Company Search**... | |
| | 4. How many manufacturers provide additive manufacturing in each state? | In Michigan, there are 25 manufacturers that provide additive manufacturing. In Oregon, there are 10 manufacturers. In Ontario, there are 17 manufacturers… | I'm unable to find the specific number of additive manufacturing companies in each state from the available resources. However, I did find a directory listing companies involved in additive manufacturing... | |
| | 5. How many manufacturers located in Michigan, provide welding but not certified with AWS? | There are 173 manufacturers located in Michigan that provide welding but are not certified with AWS. | The search did not directly yield information on manufacturers in Michigan that provide welding services but are not certified by the American Welding Society (AWS). However, there are various directories and resources that list welding service providers ... | |
| | 6. Which State has the biggest number of manufacturers which provide additive manufacturing and provide casting? | California has the biggest number of manufacturers which provide additive manufacturing and provide casting, with a total of 9 manufacturers. | Determining the U.S. state with the highest number of manufacturers offering both additive manufacturing and casting services is challenging without access to specific and current industry databases or reports. However, I can provide a general insight based on known manufacturing hubs and industry trends... | The responses from GPT-4 are based on pattern recognition and educated estimations, the terms like, "general", "likely", "might", "infer" indicate that they lack definitive reliability. In contrast, answers derived from the knowledge graph are supported by structured data distribution, offering greater |
| | 7. List Top 5 States which have the | The top 5 states with the biggest number of manufacturers that | Identifying the top five U.S. states with the most manufacturers providing injection molding services and certified with AS 9100, a quality | |
| | biggest number of manufacturers which provide injection molding and are certified with AS9100? | provide injection molding and are certified with AS9100 are California, Texas, Connecticut, Washington, and Ontario. | management standard for the aerospace industry, requires specific industry data that isn't readily available in open-source formats. However, based on general manufacturing trends and the prominence of the aerospace industry, the following states are likely candidates… | reliability and interpretability. |
| | 8. For manufacturers located in California and certified with ITAR, what service do they provide the most? | For manufacturers located in California and certified with ITAR, the service they provide the most is machining. | Determining the most provided service by ITAR-certified manufacturers in California requires specific industry data, which is not readily available in a summarized format. However, based on general knowledge of the industry and the nature of ITAR (International Traffic in Arms Regulations) certification, we can infer the types of services that are likely to be most prevalent among these manufacturers... | |
| | 9. List 40 names of manufacturers located in California which provide machining and are certified with ISO9001. | www.mossprecision.com www.iconn-ems.com www.juellmachine.com ... | I was able to identify 12 manufacturers in Southern California that are ISO 9001 certified and provide machining services. However, to list 30 such manufacturers, additional research is needed… | GPT-4's responses provide less manufacturers than required. To fully address such queries, integrating GPT-4's language abilities with the detailed data from the MSKG would yield more comprehensive and accurate results. |
| | 10. List 30 names of manufacturers which provide the same service as manufacturer 110metalworks.com. | www.tpmachinecompany.com www.coyotecutters.com klmtoolcompany.com www.preciseturning.com www.innovacera.com … | ...Unfortunately, I could only find 24 manufacturers that provide services similar to 110 Metalworks. If you need more names, I recommend further searching for metal parts and assemblies manufacturing companies. | |
| | 11. List Top 10 manufacturing services which manufacturer provides the most in North Carolina and how many manufacturers provide them. | The top 10 manufacturing services in North Carolina, along with the number of manufacturers that provide them, are as follows:\n\n1. Machining - 123 manufacturers\n2. Welding - 77 manufacturers\n3. Assembly - 75 manufacturers… | I found information on the top manufacturing services in North Carolina, but I was unable to find specific numbers for how many manufacturers provide each service… | The answers from GPT-4 are incomplete which only address the first part of questions with general assumption while the other half of questions exceed the capability to solve. However, by using knowledge graphs, not only the first part of questions is effectively addressed with graph data support, but also |
| | 12. List services provided by manufacturers certified with ITAR and located in California and how many manufacturers provide them. | turning 11, milling 9, inspection 14, grinding 5 … | In Southern California, there are 18 manufacturers that are ITAR registered and provide additive manufacturing services. Services provided by these companies range widely, including but not limited to: Custom distribution and supply chain solutions, engineering support, repair and inspection, and inventory management... | the second part is solved using complex queries over the graph. Overall, the answers using MSKG are more reliable, complete, and interpretable. |
| Hard (Graph Embedding -based) | 13. Give me 300 manufacturers similar to 110metalworks.com based on the services they provide. | 110metalworks.com alliedplastics.com www.jonesfabmachine.com www.advancedmfgservice.com modelcraft.net … | Generating a list of 300 manufacturers similar to 110 Metalworks, which specializes in custom metal fabrication, welding, and machining services, is quite extensive and would require detailed industry research. However, I can guide you on how to find similar manufacturers and what criteria to use when searching for them... | The questions exceed GPT-4's capabilities alone but can be effectively answered by combining GPT-4 with graph embedding results from MSKG. |
| | 14. Give me 300 manufacturers similar to 3d-cam.com based on the services they provide. | 3d-cam.com www.solmet.net pkerengineering.com absolutedsi.com ... | Compiling a list of 300 manufacturers similar to 3D-CAM, which specializes in rapid prototyping, 3D printing, and injection molding services, is a substantial task that requires detailed industry research. However, I can guide you on how to find similar manufacturers and the criteria you might use:1. **Industry Directories** 2. **Trade Associations**... | |
| | 15. Label "3d-cam.com" with the following tags: 1-machining, 2-assembly, 3-joining, 4-inspection, 5-forming, 6-molding, 7-casting, 8-additive manufacturing, 9-heat treatment and 10-other? | 3d-cam.com: 1-machining, 6-molding, 7-casting, 8-additive manufacturing | … Here's how each could be labeled based on typical services they might provide- **1-Machining**: Likely, as they offer CNC machining services. - **2-Assembly**: Possible… | GPT-4's response relies on generalized predictions, whereas the MSKG's answers, derived from graph embeddings, are being evaluated for their high reliability. |

[17] Gong, F., Wang, M., Wang, H., Wang, S., & Liu, M. (2021). SMR: Medical KMOWLEDGE GRAPH Embedding for Safe Medicine Recommendation. Big Data Research, 23, 100174. https://doi.org/10.1016/j.bdr.2020.100174

[18] Wang, Q., Li, M., Wang, X., Parulian, N., Han, G., Ma, J., Tu, J., Lin, Y., Zhang, H., Liu, W., Chauhan, A., Guan, Y., Li, B., Li, R., Song, X., Fung, Y. R., Ji, H., Han, J., Chang, S.-F., … Onyshkevych, B. (2021). COVID-19 Literature KMOWLEDGE GRAPH Construction and Drug Repurposing Report Generation. ArXiv:2007.00576 [Cs]. http://arxiv.org/abs/2007.00576

[19] Zhao, M., Wang, H., Guo, J., Liu, D., Xie, C., Liu, Q., & Cheng, Z. (2019). Construction of an Industrial KMOWLEDGE GRAPH for Unstructured Chinese Text Learning. Applied Sciences, 9(13), 2720. https://doi.org/10.3390/app9132720

[20] Nguyen, H., & Patrick, J. (2016). Text Mining in Clinical Domain: Dealing with Noise. Proceedings of the 22nd ACM SIGKDD International Conference on Knowledge Discovery and Data Mining, 549–558. https://doi.org/10.1145/2939672.2939720

[21] Devlin, J., Chang, M.-W., Lee, K., & Toutanova, K. (2019). BERT: Pre-training of Deep Bidirectional Transformers for Language Understanding. ArXiv:1810.04805 [Cs]. http://arxiv.org/abs/1810.04805

[22] Bubeck, S., Chandrasekaran, V., Eldan, R., Gehrke, J., Horvitz, E., Kamar, E., Lee, P., Lee, Y. T., Li, Y., Lundberg, S., Nori, H., Palangi, H., Ribeiro, M. T., & Zhang, Y. (2023). Sparks of Artificial General Intelligence: Early experiments with GPT-4. https://doi.org/10.48550/ARXIV.2303.12712

[23] Mulyar, A., & McInnes, B. T. (2020). MT-Clinical BERT: Scaling Clinical Information Extraction with Multitask Learning. ArXiv:2004.10220 [Cs]. http://arxiv.org/abs/2004.10220

[24] Luo, Y., Wu, W., & Lyu, P. (2021). Single Information Extraction Algorithm of Mechanical Equipment Usage Information Recording Based on Deep Learning. Journal of Physics: Conference Series, 1865(4), 042032. https://doi.org/10.1088/1742-6596/1865/4/042032

[25] Lewis, M., Liu, Y., Goyal, N., Ghazvininejad, M., Mohamed, A., Levy, O., Stoyanov, V., & Zettlemoyer, L. (2019). BART: Denoising Sequence-to-Sequence Pre-training for Natural Language Generation, Translation, and Comprehension. https://doi.org/10.48550/ARXIV.1910.13461

[26] Chase, H. (2023). Langchain. Retrieved July 17, 2023, from https://github.com/hwchase17/langchain

[27] Liu, J. (2022). LlamaIndex [Software]. Zenodo. https://doi.org/10.5281/zenodo.1234

[28] Karray, M. H., Otte, N., Kiritsis, D., Ameri, F., Kulvatunyou, B., Will, C., Arista, R., Rai, R., & Smith, B. (n.d.). The Industrial Ontologies Foundry (IOF) perspectives. 6.

[29] Lu, Y., Wang, H., & Xu, X. (2019). ManuService ontology: A product data model for service-oriented business interactions in a cloud manufacturing environment. Journal of Intelligent Manufacturing, 30(1), 317–334.

[30] Yan, H., Yang, J., & Wan, J. (2020). KnowIME: A System to Construct a KMOWLEDGE GRAPH for Intelligent Manufacturing Equipment. IEEE Access, 8, 41805–41813.

[31] He, L., & Jiang, P. (2019). Manufacturing KMOWLEDGE GRAPH: A Connectivism to Answer Production Problems Query With Knowledge Reuse. IEEE Access, 7, 101231–101244. https://doi.org/10.1109/ACCESS.2019.2931361

[32] Siddharth, L., Blessing, L. T. M., Wood, K. L., & Luo, J. (2022). Engineering KMOWLEDGE GRAPH From Patent Database. Journal of Computing and Information Science in Engineering, 22(2), 021008. https://doi.org/10.1115/1.4052293
25